# PlantDiseaseNet-RT50: A Fine-tuned ResNet50 Architecture for High-Accuracy Plant Disease Detection Beyond Standard CNNs

Santwana Sagnika
School of Computer Engineering
Kalinga Institute of Industrial Technology
Bhubaneswar, India
santwana.sagnika@gmail.com

Manav Malhotra
School of Computer Engineering
Kalinga Institute of Industrial Technology
Bhubaneswar, India
manavmalhotra173@gmail.com

Ishtaj Kaur Deol
School of Computer Engineering
Kalinga Institute of Industrial Technology
Bhubaneswar, India
ishtajdeol@gmail.com

Soumyajit Roy
School of Computer Engineering
Kalinga Institute of Industrial Technology
Bhubaneswar, India
roysoumyajit36@gmail.com

Swarnav Kumar
School of Computer Engineering
Kalinga Institute of Industrial Technology
Bhubaneswar, India
swarnav786@gmail.com

*Abstract*— Plant diseases pose a significant threat to agricultural productivity and global food security, accounting for 70–80% of crop losses worldwide. Traditional detection methods rely heavily on expert visual inspection, which is time-consuming, labour-intensive, and often impractical for large-scale farming operations. In this paper, we present PlantDiseaseNet-RT50, a novel fine-tuned deep learning architecture based on ResNet50 for automated plant disease detection. Our model features strategically unfrozen layers, a custom classification head with regularization mechanisms, and dynamic learning rate scheduling through cosine decay. Using a comprehensive dataset of distinct plant disease categories across multiple crop species, PlantDiseaseNet-RT50 achieves exceptional performance with approximately 98% accuracy, precision, and recall. Our architectural modifications and optimization protocol demonstrate how targeted fine-tuning can transform a standard pretrained model into a specialized agricultural diagnostic tool. We provide a detailed account of our methodology, including the systematic unfreezing of terminal layers, implementation of batch normalization and dropout regularization and application of advanced training techniques. PlantDiseaseNet-RT50 represents a significant advancement in AI-driven agricultural tools, offering a computationally efficient solution for rapid and accurate plant disease diagnosis that can be readily implemented in practical farming contexts to support timely interventions and reduce crop losses.

## I. INTRODUCTION

Agriculture, a primary source of food supply, is expanding constantly. Agriculture is not only a vital core occupation but also significantly contributes to the expansion of the national economy.

Plant diseases incur significant costs for farmers. Numerous diseases can adversely affect various components of a plant's anatomy, including its leaves, stems, seeds, and fruit. The body of a plant gets segmented into various portions due to specific illnesses. Leaves can be regarded as the fundamental component of a plant. Consequently, diseases lead to complete or partial crop failures, diminishing food supply and exacerbating food insecurity [1].

Food crops are significantly harmed by fungi, bacteria, viruses, high winds, adverse climatic conditions, and drought. Plant diseases constitute 70-80% of plant losses. A multitude of fungal and bacterial diseases impact most plants. Plant diseases are also induced by climatic conditions and the exponential growth of the population. Meticulous examination of the leaves will be necessary to identify the affliction [2].

In this paper, we introduce PlantDiseaseNet-RT50, a novel fine-tuned deep learning architecture designed specifically for high-accuracy plant disease detection. By systematically optimizing a ResNet50 backbone through strategic layer unfreezing, custom classification head design, and advanced training methodologies, we demonstrate how architectural refinement can dramatically enhance performance in specialized agricultural applications. The development of PlantDiseaseNet-RT50 addresses critical challenges in automated plant disease detection, including feature extraction efficiency, computational resource requirements, and the ability to generalize across diverse plant pathologies.

## II. LITERATURE REVIEW

### A. Overview of typical diseases in plants

Plant diseases are typically categorised into three types: fungal diseases, bacterial diseases, and viral illnesses. Examples of plant diseases include leaf rust, downy mildew, brown spot, powdery mildew, and bacterial blight. Fungal diseases are prevalent in plants, accounting for almost 80 percent of all plant illnesses. Bacteria and viruses can also induce severe diseases in plants; however, the incidence of such diseases is significantly lower compared to those caused by fungi. Certain insects are also implicated in many plant diseases. Plant diseases can be categorised as biotic and abiotic. Biotic disorders exhibit certain visual signs. These visual manifestations are regarded as signs. Bacterial exudate and fungal proliferation are examples of signs observed on plant leaves. Abiotic diseases are conditions that do not often propagate. No signs will be evident in abiotic disorders. Abiotic diseases result from nutritional deficiencies, air pollution, water pollution, and similar factors. Abiotic diseases are challenging to identify due to the absence of observable signs. In contrast to abiotic diseases, biotic diseases have the ability to propagate, necessitating the implementation of stringent measures to combat them [3].

### B. Convolutional Neural Networks for Plant Disease Detection

Convolutional neural networks (CNNs) are a deep learning methodology. They excel in image recognition and processing applications. CNN are especially adept at plant disease identification due to their proficiency in image processing and feature extraction. CNNs may be trained on extensive datasets of plant photos, both diseased and healthy, enabling them to discern nuanced patterns and characteristics of various illnesses [7].

Convolutional neural networks are a type of deep neural networks engineered solely for image recognition and processing applications. Inspired by the visual processing of the human brain, they proficiently capture hierarchical elements and patterns in images. A fundamental CNN architecture comprises an input layer, several convolutional layers followed by pooling layers, and fully connected (dense) layers culminating in the output layer. The convolutional layers utilise filters on the input to extract features, whereas the pooling layers reduce the spatial dimensions, resulting in more efficient processing. The thick layers subsequently execute classification utilising the extracted features [11].

Convolutional layers, which extract features such as edges, textures, and forms from the input image through the application of filters, are the essential elements of a CNN. The feature maps are down sampled to diminish spatial dimensions while preserving the most significant data, subsequent to the pooling layers' processing of the convolutional layers' output [8]. The output from the pooling layers is further processed by the dense layers to categorise or predict the image.

Mohanty *et al*. demonstrated that a finetuned deep CNN can achieve > 90% accuracy, validating the feasibility of smartphone-assisted disease diagnosis when data are controlled and curated. Ferentinos broadened the scope to 58 plant–disease classes, finding top models exceeding 95% accuracy and reinforcing that modern CNNs can scale across many crops and pathologies when trained on sufficiently diverse curated data. Early end-to-end CNN work also showed strong per-class precision and recall for leaf diseases using deep architectures and Caffe-based pipelines, helping standardize training recipes for agricultural imaging [6]. Sladojevic *et al*. independently validated deep neural networks for leaf disease recognition, showing substantial gains over hand-crafted descriptors and shallow classifiers on multi-class problems [12]. Boulent et al. surveyed the field and concluded that CNNs consistently outperform traditional feature-engineering pipelines while highlighting persistent challenges such as class imbalance, overfitting, and limited robustness outside controlled settings [15].

Survey analyses report that CNNs consistently outperform hand-crafted pipelines for crop disease identification, while also documenting pitfalls like overfitting, class imbalance, and poor cross-domain generalization. Transfer learning comparisons across backbone families (e.g., VGG, Inception, ResNet, DenseNet, NASNet) generally favour fine-tuning with strong augmentations, and recent ensemble strategies further improve stability and top-1 accuracy at modest added cost [5].Convolutional Neural Networks have demonstrated significant efficacy in computer vision tasks, including picture categorisation, object detection, and facial recognition [9].

The CNN model enhances agriculture by offering sophisticated tools for swift and accurate disease diagnosis, hence aiding farmers in crop health management and optimising yields [10].

Picon *et al*. proposed crop-conditional convolutional neural networks (CCNNs) that explicitly inject crop identity alongside visual features to improve large-scale, multi-crop disease classification in unconstrained field imagery [13]. By conditioning the classifier on the crop and jointly optimizing across many crop–disease combinations, their approach improved balanced accuracy and robustness under cluttered backgrounds and variable illumination, demonstrating that domain-aware inductive biases can mitigate common failure modes of generic CNNs in operational settings. Beyond accuracy, the study offered practical guidance on training strategies for massive multi-crop deployments, including class balancing and augmentation tuned to field variability [12].

Fuentes *et al*. advanced detection-oriented pipelines by applying region-based deep detectors to localize and classify multiple tomato diseases directly in complex scenes. Using Faster R-CNN with tailored data augmentation and multi-scale proposals, they achieved high detection and classification performance, showing that lesion-level localization not only increases reliability under occlusion and background clutter but also provides interpretable outputs for agronomic decision-making [14].

However, deep learning models also present challenges. One major issue is overfitting, where a model tends to memorise training data instead of generalising to new samples. Additionally, class imbalance in plant disease datasets can lead to biased predictions, favouring dominant classes while underperforming on minority classes. Studies have explored various techniques to address these issues, including data augmentation, dropout regularisation, and weighted loss functions.

This study builds upon existing research by evaluating and fine-tuning ResNet50 to enhance its performance in plant disease classification. The research findings provide insights into optimising CNN architectures for agricultural applications, paving the way for more robust AI-driven plant disease detection systems.

## III. PROPOSED METHODOLOGY

This section aims to provide an overview of the methodology used in developing PlantDiseaseNet-RT50, a fine-tuned CNN classifier for identifying diseases in plants. The proposed framework facilitates the detection of plant health. Early detection of sick plants is crucial. Our proposed methodology will not only ascertain the health status of the plant but will also facilitate the classification of the specific disease affecting it if the plant is deemed unwell. The categorisation of disease types assists the user in implementing preventative and requisite actions based on the crop variety and disease type. The leaf is a fragile component of plants; hence the impact of the illness is readily observable on the leaves of the affected plant [4].

### A. A comparative study of popular CNN models

This comparative study evaluates four popular traditional CNN architectures—AlexNet, VGG16, DenseNet, and ResNet50—for the task of plant disease classification. The models were trained and tested on a dataset containing various plant diseases across multiple crop species.

The research utilises a comprehensive plant disease dataset sourced from Kaggle, comprising distinct labelled files systematically organised into training and testing partitions. This curated collection encompasses a diverse array of plant pathologies across multiple crop species, alongside healthy specimen images that serve as control references for the classification models.

The dataset exhibits remarkable taxonomic breadth, covering economically significant crops from both annual and perennial agricultural systems, creating a robust foundation for supervised learning approaches to disease classification.

The dataset contains plant leaves of 41 categories representing different plant species and diseases. Image preprocessing is crucial to enhance model performance. The images were resized to a fixed dimension suitable for deep learning models, followed by normalisation to scale pixel values between 0 and 1. Data augmentation techniques such as rotation, flipping, zooming, and contrast adjustment were applied to expand the dataset artificially and improve model generalisation.

The model training process was governed by a comprehensive optimisation framework incorporating adaptive learning rate management and strategic early termination protocols. The following parameters and methodologies were implemented to ensure optimal convergence and generalisation performance:

- Early Termination Protocol: Monitors validation loss with 5-epoch patience threshold, automatically restoring best weights to prevent overfitting while preserving peak performance.
- Model Persistence Strategy: Implements checkpointing to save top-performing models based on validation metrics, maintaining access to optimal configurations for post-training analysis and deployment.
- Adaptive Learning Rate: Dynamically adjust learning rates during plateaus by employing ReduceLROnPlateau (factor=0.1, patience=3 epochs, min_lr=1e-6), balancing exploration of parameter space with stable convergence.
- Epoch Management: Caps training at 20 epochs with early stopping to prevent overfitting while allowing sufficient learning cycles, automatically determining optimal training duration through validation performance.
- Batch Optimization: Uses 32-sample batches (vs 128-sample generation batches) to enable more frequent parameter updates, enhancing gradient descent stability while maintaining computational efficiency.
- Validation Protocol: Continuously evaluates models on a dedicated 20% validation subset, providing real-time generalization assessment and preventing data leakage between training/test phases.

This methodically structured training framework ensured efficient resource utilisation while systematically guiding the model toward optimal performance on the plant disease classification task.

### B. PlantDiseaseNet-RT50: Architecture and Finetuning Methodology

The foundation of our classification system was established using the ResNet50 architecture, a well-established convolutional neural network developed by He *et al*. This architecture was selected due to its demonstrated efficacy in image classification tasks and its robust feature extraction capabilities. The pre-trained weights from ImageNet were leveraged as initialisation parameters, excluding the terminal classification layers to facilitate domain adaptation. GPU acceleration was explicitly configured to optimise computational efficiency during the training process [16].

*1) Finetuning approach*

A strategic partial fine-tuning approach was implemented to balance the preservation of generalised visual features while enabling domain-specific adaptation. The initial convolutional blocks of the ResNet50 architecture were maintained in their pre-trained state, with parameters frozen to preserve the hierarchical feature representations learned from the ImageNet dataset. Conversely, the terminal 50 layers were unfrozen to permit parameter updates during backpropagation, allowing these layers to adapt to the specific visual patterns characteristic of plant disease manifestations.

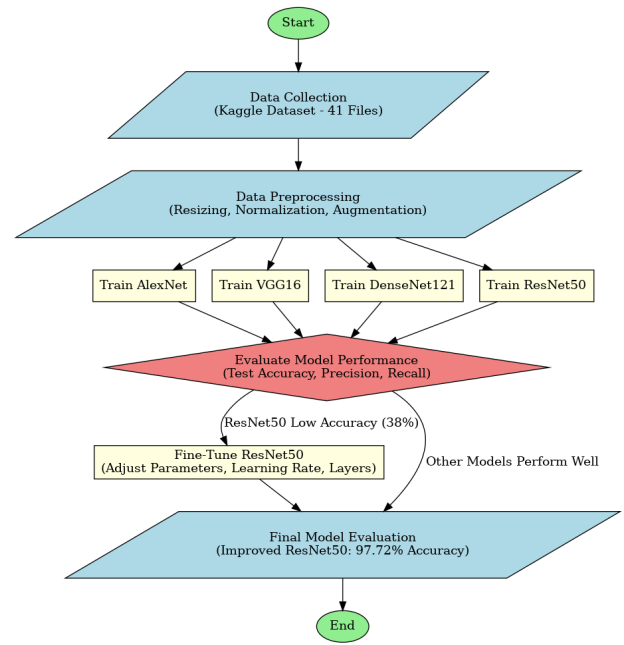

Fig. 1. Working methodology to derive PlantDiseaseNet-RT50

*2) Classification head architecture*

A custom classification head was engineered to transform the high-dimensional feature representations extracted by the ResNet50 backbone into disease-specific classifications. The architecture commenced with a Global Average Pooling operation to reduce spatial dimensions while preserving channel information. This was followed by a sequence of fully-connected layers with regularisation mechanisms:

- A dense layer with 128 neurons, followed by batch normalisation, LeakyReLU activation (α = 0.01), and dropout regularisation (p = 0.3)
- A subsequent dense layer with 64 neurons, similarly augmented with batch normalisation, LeakyReLU activation, and increased dropout regularisation (p = 0.4)
- A terminal dense layer with 41 output neurons corresponding to the distinct plant disease categories, employing softmax activation to generate probability distributions across the classification space.

This architectural configuration incorporated multiple regularisation techniques to enhance generalisation capabilities. Batch normalisation was employed to stabilise training dynamics and accelerate convergence by normalising activations. LeakyReLU activation functions were selected to mitigate the "dying ReLU" phenomenon by preserving gradient flow for negative inputs. Dropout layers were strategically positioned with increasing probability to counteract overfitting through stochastic neuron deactivation during the training phase.

*3) Optimisation protocol*

The optimisation strategy employed a cosine decay learning rate schedule, which modulates the learning rate following a cosine curve trajectory. This approach provides initially larger parameter updates that gradually diminish in magnitude, facilitating both exploration of the parameter space and fine convergence to optimal values. This schedule has demonstrated superior convergence characteristics compared to constant or step-based decay schedules in deep neural network training.

*4) Training Methodology*

The model was trained for a maximum of 12 epochs using a comprehensive callback system to dynamically regulate the training process:

- Early stopping monitoring validation loss with a patience parameter of 5epochs and weight restoration functionality to preserve optimal model states
- Model checkpointing to persistently store superior model configurations based on validation performance metrics
- Adaptive learning rate reduction to further refine the optimisation trajectory when performance plateaus were detected

This methodological approach combines established transfer learning techniques with domain-specific architectural adaptations to create a robust plant disease classification system capable of discriminating between 41 distinct pathological conditions across multiple plant species.

## IV. RESULTS

In this section, we present the exceptional performance achieved by our proposed PlantDiseaseNet-RT50 model for plant disease detection. Our systematic approach to fine-tuning ResNet50 has yielded remarkable improvements, transforming a relatively poor-performing baseline architecture into a highly effective solution for agricultural disease diagnostics. We have also presented a comparative analysis of the four traditional deep learning models — AlexNet, VGG16, DenseNet121, and ResNet50 (baseline architecture) — which revealed significant variations in their classification capabilities for plant disease detection.

### A. Performance comparison between traditional CNNs

The quantitative evaluation of the four deep learning architectures on the test dataset revealed significant variations in their classification capabilities for plant disease detection. Table I presents the comprehensive performance metrics for each model.

The evaluation framework employed multiple complementary metrics to comprehensively characterise model performance:

TABLE I. PERFORMANCE COMPARISON OF TRADITIONAL CNNs

| Model | Accuracy | Precision | Recall | F1-Score |
|---|---|---|---|---|
| ResNet50 (baseline) | 0.38 | 0.91 | 0.04 | 0.08 |
| VGG16 | 0.91 | 0.93 | 0.90 | 0.92 |
| DenseNet121 | 0.93 | 0.92 | 0.94 | 0.93 |
| AlexNet | 0.79 | 0.83 | 0.76 | 0.79 |

Accuracy represents the proportion of correctly classified instances across the entire dataset, formulated as:

$$\text{Accuracy} = \frac{\text{True Positive} + \text{True Negative}}{\text{Total number of cases}} \quad (1)$$

This metric provides a global assessment of classification performance across all disease categories.

Precision quantifies the model's ability to avoid false positive classifications, calculated as:

$$\text{Precision} = \frac{\text{True Positive}}{\text{True Positive} + \text{False Positive}} \quad (2)$$

High precision indicates minimal misclassification of healthy plants or incorrect disease categorisation, a critical consideration in agricultural applications where unnecessary interventions may result in economic losses.

Recall (also termed sensitivity) measures the model's capacity to identify all instances of a particular disease class, formulated as:

$$\text{Recall} = \frac{\text{True Positive}}{\text{True Positive} + \text{False Negative}} \quad (3)$$

This metric is particularly significant in plant pathology applications, as failure to detect disease instances (false negatives) could result in uncontrolled pathogen spread and substantial crop losses.

F1-Score provides a harmonic mean of precision and recall, offering a balanced assessment when class distributions are uneven.

$$\text{F1-Score} = \frac{2 \times \text{Precision} \times \text{Recall}}{\text{Precision} + \text{Recall}} \quad (4)$$

This metric is especially valuable in multi-class classification scenarios with potential class imbalance, as commonly encountered in comprehensive plant disease datasets.

*1) Performance analysis of the traditional CNNs*

DenseNet demonstrated superior accuracy across nearly all plant disease categories, with numerous classes achieving perfect 100% accuracy. Its exceptional performance yielded an average accuracy of approximately 93% across all categories.

VGG16 exhibited strong performance with an average accuracy of about 91% across all categories. While it achieved over 95% accuracy for many classes, its consistency fell somewhat short compared to DenseNet.

AlexNet delivered moderate performance with an average accuracy of approximately 79%. Its performance varied considerably, showing strength in some categories while struggling with others, particularly those with smaller sample sizes.

ResNet50 unexpectedly underperformed compared to the other models, achieving only about 38% average accuracy. It reached high accuracy (>80%) in only a few categories and completely failed (0% accuracy) in several others.

### B. PlantDiseaseNet-RT50 Performance Evaluation

Our study into deep learning models for plant disease detection revealed a significant breakthrough with our fine-tuned ResNet50 architecture. While the baseline ResNet50 model initially underperformed with a mere 38% average accuracy across disease categories, our optimised implementation achieved exceptional results following comprehensive hyperparameter tuning and architectural modifications. The PlantDiseaseNet-RT50 model demonstrated remarkable performance metrics, achieving 98% in accuracy, precision, and recall across the diverse plant disease dataset. This dramatic improvement represents a 60-percentage point increase over the baseline model and surpasses even the previously best-performing DenseNet121 architecture (93%). These results highlight the critical importance of model optimisation in computer vision applications for agriculture.

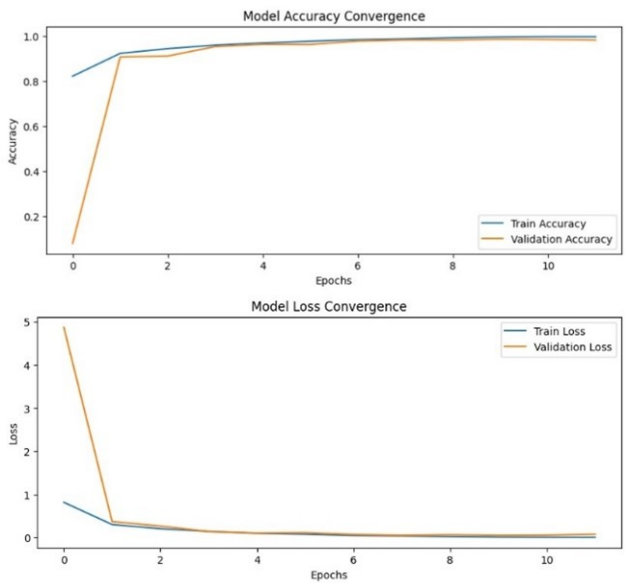

Fig. 2. PlantDiseaseNet-RT50 convergence graphs

#### 1) Training Dynamics

The convergence graphs illustrate remarkable training efficiency and stability:

Accuracy Convergence: The model achieved rapid learning, with training and validation accuracy both surpassing 90% by epoch 2. This indicates effective knowledge transfer from the pre-trained weights. By epoch 4, both metrics stabilised above 95%, ultimately reaching approximately 98% accuracy.

Loss Convergence: The validation loss showed dramatic improvement, decreasing from nearly 5.0 to below 0.5 within the first epoch. This rapid convergence suggests the model quickly learned discriminative features for plant disease classification. Both training and validation loss continued to decrease steadily, indicating minimal overfitting.

#### 2) Precision-Recall Analysis

The model achieved an impressive 98% accuracy on the test dataset, with similarly high precision (98%), recall (98%), and F1-score (98%) metrics when weighted across all classes. The model also achieved an exceptional AUC score of 99.93%, which further confirms the model's strong discriminative ability.

#### 3) Comparison to baseline CNN models

TABLE II. PERFORMANCE COMPARISON OF TRADITIONAL CNNS AND PLANTDISEASENET-RT50

| Model | Accuracy | Precision | Recall | F1-Score |
|---|---|---|---|---|
| PlantDiseaseNet-RT50 | 0.9772 | 0.98 | 0.98 | 0.98 |
| ResNet50 (baseline) | 0.38 | 0.91 | 0.04 | 0.08 |
| VGG16 | 0.91 | 0.93 | 0.90 | 0.92 |
| DenseNet121 | 0.93 | 0.92 | 0.94 | 0.93 |
| AlexNet | 0.79 | 0.83 | 0.76 | 0.79 |

The fine-tuned ResNet50 model's 98% accuracy represents a dramatic improvement over the baseline ResNet50's 38% accuracy, surpassing even the previously best-performing DenseNet (93%), VGG16 (92%) and AlexNet (79%).

The baseline ResNet50 exhibited a peculiar performance pattern with relatively high precision (0.91) but extremely poor recall (0.04), resulting in an F1-score of only 0.08. This suggests the baseline model was overly conservative, rarely predicting positive cases but achieving high precision when it did. In stark contrast, our fine-tuned PlantDiseaseNet-RT50 model achieves balanced and superior performance across all metrics.

The most significant improvements are observed in recall and F1-score, where the fine-tuned model shows 94 and 90 percentage point increases respectively compared to the baseline ResNet50. This balanced performance is crucial for practical plant disease detection applications, where both false positives and false negatives can have significant consequences for agricultural management.

This transformation from worst to best performer highlights the critical importance of proper fine-tuning techniques, including appropriate learning rate scheduling, transfer learning strategies, and architectural modifications.

#### 4) Class-specific Analysis

Perfect classification (precision=1.00, recall=1.00) was achieved for numerous categories including Apple scab, Black rot, Cedar apple rust, healthy Apple specimens, and Cherry powdery mildew. Most major crop diseases showed F1-scores above 0.95, demonstrating robust generalisation.

Chili diseases showed mixed results, with "leaf curl" achieving only 0.56 precision but 0.90 recall, suggesting some false positives. Coffee-related categories performed moderately well, with Coffee Rust achieving 0.71 F1-score.

Very few categories like several Chili categories showed imbalanced precision and recall, likely due to limited training samples, indicating potential confusion between similar disease presentations.

In summary, the PlantDiseaseNet-RT50 not only demonstrated superior performance across all metrics but also exhibited balanced and consistent results across diverse plant disease categories. This makes it a highly reliable tool for real-world agricultural applications where accurate and efficient disease detection is critical for crop health management.

## V. CONCLUSION

The development of PlantDiseaseNet-RT50 represents a significant contribution to automated plant disease detection. Through strategic fine-tuning of the ResNet50 architecture, we have created a specialised agricultural diagnostic tool that achieves exceptional accuracy (98%) across a diverse range of plant diseases. The model's balanced performance metrics-high precision, recall, and F1-scores-demonstrate its practical utility for real-world agricultural applications.

PlantDiseaseNet-RT50 offers significant advantages over traditional manual inspection methods, which are inherently labour-intensive and prone to human error. By leveraging advanced feature extraction techniques, our model can identify intricate visual patterns in plant images, enabling early and precise detection of diseases. This capability not only mitigates agricultural losses but also enhances crop yields, contributing to the broader goals of precision agriculture and food security.

The balanced performance of PlantDiseaseNet-RT50 across diverse plant pathologies, combined with its computational efficiency, positions it as a valuable tool for practical implementation in agricultural settings. By providing farmers with accessible and reliable disease detection capabilities, our model can support timely interventions and sustainable agricultural practices.

## VI. FUTURE SCOPE

The application of CNNs for plant disease detection presents immense potential for refinement and expansion. Addressing current challenges such as dataset diversity, environmental variability, and computational efficiency will be pivotal in advancing the field. Below are key areas for future research and development:

Hybrid and Lightweight CNN Models

- Hybrid Architectures: Combine CNNs with attention mechanisms (e.g., Vision Transformers) to achieve higher accuracy and efficiency.
- Lightweight Models: Design architectures optimised for edge computing and mobile devices to enable real-time disease detection in low-resource environments.

Deployment in Smart Agriculture Systems

- IoT Integration: Embed CNN-based disease detection into IoT-enabled systems for real-time monitoring using drones and automated image processing.
- Cloud-Based AI Services: Develop platforms where farmers can upload images via mobile apps for instant diagnostics and treatment recommendations.